%% file: verb-main.tex
\documentclass[journal]{vgtc-no-name}

\ifpdf                                
  \pdfoutput=1\relax                   
  \usepackage{graphicx}                
  \DeclareGraphicsExtensions{.pdf,.png,.jpg,.jpeg} 
  \else                                 
  \usepackage{graphicx}                
  \DeclareGraphicsExtensions{.eps}     
\fi

\graphicspath{{figs/}{./}}

\usepackage{microtype}                 
\PassOptionsToPackage{warn}{textcomp}  
\usepackage{textcomp}                  
\usepackage{mathptmx}                  
\usepackage{times}                     
         
\usepackage{cite}                      
\usepackage{tabu}                      
\usepackage{booktabs}                  
\usepackage{algorithmic}
\usepackage{algorithm}
\usepackage{wrapfig}

\usepackage{amsfonts}

\newcommand {\mm}[1] {\ifmmode{#1}\else{\mbox{\(#1\)}}\fi}
\newcommand{\Rspace}        {\mm{\mathbb{R}}}

\newcommand{\para}[1]  {\vspace{1mm}\noindent{{\textbf{#1}}}}

\newcommand{\etal}{\textit{et al.}}
\newcommand{\wrt}{\textit{w.r.t.}}

\newcommand{\tool} {\mbox{VERB}}
\newcommand{\denselist}{\vspace{-5pt} \itemsep -3pt\parsep=-1pt\partopsep -3pt}

\usepackage[dvipsnames]{xcolor}

\onlineid{0}

\vgtccategory{Research}
\vgtcpapertype{Analytics \& Decisions}

\title{VERB: Visualizing and Interpreting Bias Mitigation Techniques 
\\for Word Representations}

\author{Archit Rathore, Sunipa Dev, Jeff M. Phillips, Vivek Srikumar, \\
Yan Zheng, Chin-Chia Michael Yeh, Junpeng Wang, Wei Zhang, Bei Wang}

\authorfooter{
\item Archit Rathore, Jeff M. Phillips, Vivek Srikumar, and Bei Wang are with the University of Utah. E-mails: archit.rathore@utah.edu, jeffp@cs.utah.edu, svivek@cs.utah.edu, beiwang@sci.utah.edu.
\item Sunipa Dev is with the University of California, Los Angeles. E-mail: sunipa@cs.ucla.edu.
\item Yan Zheng, Michael Yeh, Junpeng Wang, and Wei Zhang are with VISA research. E-mails: \{yazheng, miyeh, junpenwa, wzhan\}@visa.com.
}

\shortauthortitle{Rathore \MakeLowercase{\textit{et al.}}: VERB}

\abstract{
\input{sec-abstract}
}


\teaser{
  \centering
  \includegraphics[width=1.0\linewidth]{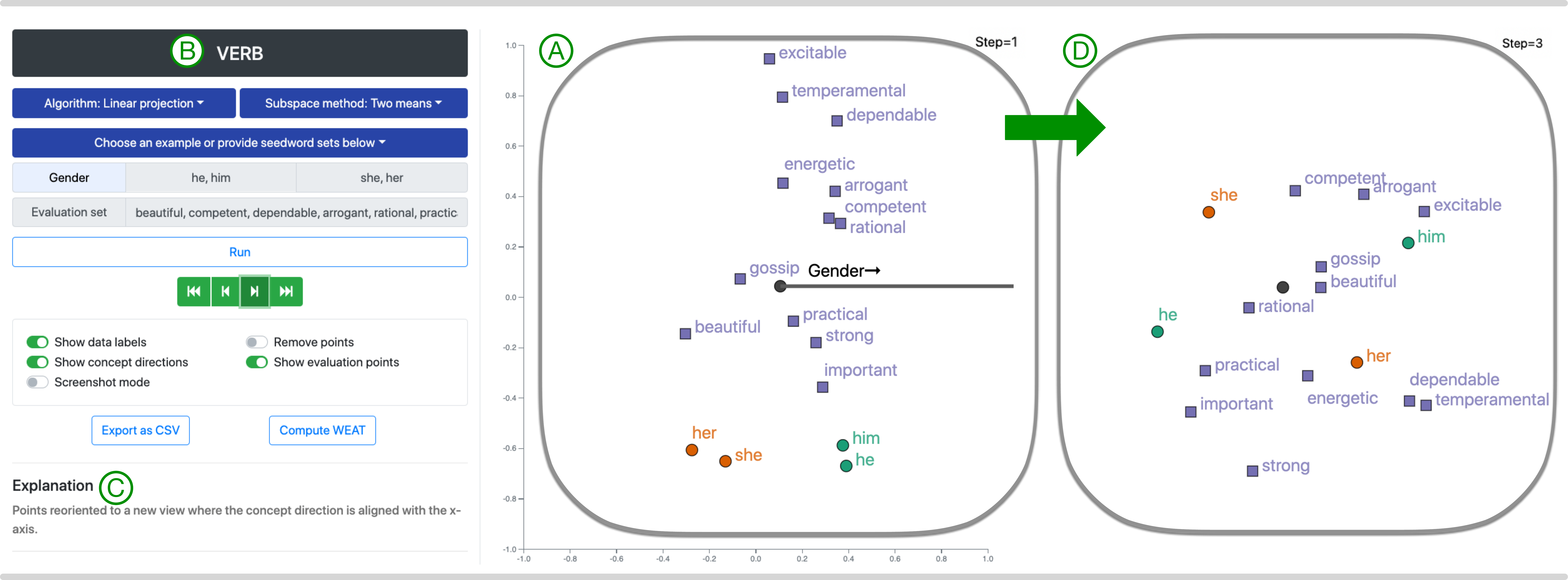}
   \caption{With VERB, users can explore the high-dimensional word representations interactively before, during, and after applying bias mitigation techniques. (A) \textbf{Embedding View} highlights a subset of word embeddings using dimensionality reduction and visualizes the step-by-step transformations of their embeddings across various debiasing techniques. (B) \textbf{Control Panel} enables users to configure each debiasing technique and provides controls to iterate through each step of the transformation. (C) \textbf{Explanation Panel} gives a step-by-step description of the transformation. In this example, in the Embedding View, word representations are reoriented with the concept direction (Gender) along the x-axis.  They are then transformed by a debiasing technique called the Linear Projection that removes this concept. These transformed representations are visualized again  in (D) after a new dimensionality reduction, where there is no longer a well-defined gender concept, and hence the stereotypical associations are mitigated.}  
\label{fig:teaser}
}

\nocopyrightspace
\vgtcinsertpkg

\begin{document}
\maketitle

\input{sec-introduction.tex}

\input{sec-related-work.tex}
\input{sec-debiasing.tex}
\input{sec-tool.tex}
\input{sec-use-cases.tex}
\input{sec-feedback.tex}

\input{sec-conclusions.tex}

\section*{Acknowledgments}
This project was partially supported by NSF DBI-1661375,  IIS-1816149, CCF-1350888, and a grant from VISA Research.


\input{verb-main.bbl}
\end{document}

%% file: sec-abstract.tex
Word vector embeddings have been shown to contain and amplify biases in data they are extracted from.
Consequently, many techniques have been proposed to identify, mitigate, and attenuate these biases in word representations.
In this paper, we utilize interactive visualization to increase the interpretability and accessibility of a collection of state-of-the-art debiasing techniques.
To aid this, we present Visualization of Embedding Representations for deBiasing system (``VERB"), an open-source web-based visualization tool that helps the users gain a technical understanding and visual intuition of the inner workings of debiasing techniques, with a focus on their geometric properties.
In particular, VERB offers easy-to-follow use cases in exploring the effects of these debiasing techniques on the geometry of high-dimensional word vectors.
To help understand how various debiasing techniques change the underlying geometry, VERB decomposes each technique into interpretable sequences of primitive transformations and highlights their effect on the word vectors using dimensionality reduction and interactive visual exploration.
VERB is designed to target natural language processing (NLP) practitioners who are designing decision-making systems on top of word embeddings, and also researchers working with fairness and ethics of machine learning systems in NLP.
It can also serve as a visual medium for education, which helps an NLP novice to understand and mitigate biases in word embeddings.

%% file: sec-introduction.tex
\section{Introduction}
\label{sec:introduction}

Complicated and massive data sets are becoming more and more commonly represented as vectorized embeddings.  
They are part of a de facto model for words in word vector embeddings like \emph{Word2Vec}~\cite{MikolovSutskeverChen2013} and \emph{GloVe}~\cite{PenningtonSocherManning2014}, and are becoming commonplace for other data types like graphs, spatial regions, and merchants.  
These vectorized embeddings (referred to as \emph{representations}) directly capture similarity between objects. Additional structures arise implicitly from these representations, such as linear subspaces that capture concepts (e.g., gender,  occupation, and nationality) among word vectors.  
Moreover, these representations permit easy integration into machine learning tasks.  

A downside of these representations is that their high-dimensional nature obscures easy interpretation -- at least not without much additional effort.
Although these representations do not have explicit agendas, they can nevertheless encode \emph{biases} through data imbalance and other more hidden factors.
We must emphasize that bias is a complex concept whose interpretation is being studied. In this paper, it refers to the associations of a stereotypical nature that are expressed in word representations.
For fair machine learning (ML), it is often necessary to \emph{modify} these vectorized representations to mitigate such biases.  This notably includes attenuating bias~\cite{DevPhillips2019,BolukbasiChangZou2016}, but could be other forms of normalization or alignment tasks.  

Visualization tools for these high-dimensional vector representations exist, and are overviewed in~\autoref{sec:related-work}.
However, these previous approaches are all \emph{passive}, and only allow a user to inspect such data, but not modify it.
In contrast, the tool we present -- \emph{VERB} (Visualization of Embedding Representations for deBiasing) -- is \emph{active}, and it allows a user to modify the embedding while visually exploring it.
That is, \emph{when you are using VERB, you are doing something.}  

In particular, our new tool VERB allows for an easy understanding and use of methods to debias word vector embeddings.  This allows for easy interpretation and comparison of the various methods now available.
For instance, a user can easily observe the difference between the original \emph{Hard Debiasing} approach~\cite{BolukbasiChangZou2016} or the new and simpler \emph{Linear Projection} method~\cite{DevPhillips2019}.
Another often overlooked aspect is the comparison of how concept subspaces are found: there are several methods based on PCA, or derived from clustering or classification.
In fact, by inspecting the surprising differences between these approaches, we devise a new interactive approach towards subspace identification that outperforms all of these prior methods.
VERB allows one to insert a human in the loop of this process to dynamically improve the result, and verify it is doing what is intended.

Finally, we note that VERB is applicable beyond word vector embeddings.
Any vectorized or embedded representations can easily be loaded, inspected, actively modified, and outputted.  We demonstrate this with a new application of analyzing merchant association from a global payment company.
This provides new insights which are nebulous or hidden before the use of VERB.  VERB is open source, available at \url{https://github.com/tdavislab/verb}.

\textbf{Trigger Warning:  This paper contains examples of biases and stereotypes seen in society and in language representations. These examples may be potentially triggering and offensive. The inclusion of these examples are meant to bring light to and mitigate these biases, and it is not an endorsement.}

%% file: sec-related-work.tex
\section{Related Work}
\label{sec:related-work}

\para{Visual analytics for machine learning models.}
With the recent success of machine learning (ML), a growing number of visual analytics works have also been proposed for the interpretation of ML models~\cite{arrieta2020explainable,ribeiro2016should,liu2016towards, pezzotti2017deepeyes, wongsuphasawat2017visualizing, choo2018visual, yuan2020survey, liu2017towards}.
Based on the analysis focus, these works can roughly be categorized into three groups.
The first group concentrates on the input data of ML models to better understand the data distribution~\cite{chen2020oodanalyzer, yang2020diagnosing} or to better select the high-dimensional data features~\cite{krause2014infuse, may2011guiding}. 
The second group focuses on the  intermediate data representations from ML models to interpret how the data has been transformed internally.
For example, most of the white-box interpretation solutions for deep learning models~\cite{springenberg2014striving,rauber2016visualizing,wang2018ganviz,RathoreChalapathiPalande2021} visualize the activation of different neurons from hidden layers, to reveal what have been captured/memorized by deep neural networks. 
The last group targets the output from ML models to evaluate and compare different models.
For example, \emph{ModelTracker}~\cite{amershi2015modeltracker} and \emph{Squares}~\cite{ren2016squares} use glyphs to encode the prediction probability of ML models and empower ML designers with instance-level data inspections and analysis. 
\emph{MLCube}~\cite{KahngFangChau2016} allows users to compare ML models' performances (accuracies) over subsets defined using feature conditions. 
\emph{Facets}~\cite{Facets} provides visualizations that aid in understanding ML datasets via individual feature exploration and subdivision of large data sets. 
To better disclose ML models' performance evolution, multiple visual designs for temporal confusion matrix have also been proposed, e.g.~\cite{LiuXiaoLiu2017, hinterreiter2020confusionflow}.

Our work fits well with this (third) categorization and we focus on analyzing and comparing the output from multiple ML models, in particular, debiasing techniques for word representations. 

\para{Visualization for NLP.}
Visualization has been employed for various NLP tasks such as topic modeling and sentiment analysis.  
For topic modeling, Chuang \etal~\cite{ChuangManningHeer2012} introduced \emph{Termite} to visually assess topic model quality.   
Smith \etal~\cite{SmithHawesMyers2014} presented \emph{Hi{\'e}rarchie} that interactively visualizes large, hierarchical topic models.
Liu \etal~\cite{LiuYinWang2016} created visual exploration to help users understand hierarchical topic evolution in text streams. 
For sentiment analysis, Smith \etal~\cite{SmithChuangHu2014} further presented a so-called \emph{relationship enriched visualization} that helps users explore topic models via corrections among words and topics. 
Wang \etal~\cite{WangXiaLiu2013} introduced \emph{SentiView} to analyze and visualize public sentiments of social media texts and their evolution. 
Liu \etal~\cite{LiuYuWei2013} used optimization to design \emph{StoryFlow}, a storyline visualization system that illustrates the dynamic relations among entities in a given story. 
Liu \etal~\cite{LiuLiLi2019} introduced \emph{NLIZE}, a visual analytic system that enables perturbation-driven exploration~\cite{LiuLiLiu2018} of a natural language inference model, where a user can perturb a model's input, attention, and prediction. 

Instead of focusing on abstract concepts such as topics and sentiments, our work aims to understand fine-level details presented in word embeddings, in particular, how the word embeddings are changed geometrically by various debiasing techniques. 
Word embeddings are considered as a type of word representations that allows words with similar meaning to have similar representations.
Specifically, a word vector is a high-dimensional real-valued vector where semantically similar words have similar vectors. 
Several works in the literature are most relevant to ours in terms of visually exploring the space of word embeddings, see \cite{HeimerlGleicher2018} for a survey of using visualization for interpreting word embeddings. 
Liu \etal~\cite{LiuBremerThiagarajan2018} studied the pair-wise analogy relationships of word embeddings and proposed a new projection method to better preserve the analogy relationships in the projection space. 
Rathore \etal~\cite{RathoreChalapathiPalande2021} visualized and investigated a graph-based summary of a collection of word embeddings obtained from the BERT (Bidirectional Encoder Representations from Transformers)  family  of  models~\cite{DevlinChangLee2019}. 
Our tool is similar to~\cite{LiuBremerThiagarajan2018}, in that it enables the interrogation and interpretation via projected views of embeddings, but goes beyond in guiding the modification of the embeddings. 

\para{Visualizing embeddings or latent spaces.} 
Word embeddings are a type of point cloud data where generic high-dimensional visualization techniques may be applicable (see~\cite{LiuMaljovecWang2017} for a survey).
Since word embeddings are typically obtained via neural networks, techniques developed for visualizing latent spaces or hidden representations of neural networks are also relevant. 

To visualize high-dimensional embeddings, dimensionality reduction algorithms (e.g., PCA, t-SNE~\cite{MaatenHinton2008}, and UMAP~\cite{McInnesHealySaul2018,McInnesHealyMelville2018}) are commonly used in their analysis and visualization. 
Openly available toolkits such as \emph{scikit-learn}~\cite{PedregosaVaroquauxGramfort2011} implements a number of such algorithms.
Two analytical tasks are often the focus. The \textit{first} one is to interpret the semantics encoded in embeddings (e.g.~\cite{LiuBremerThiagarajan2018}).
For example, Smilkov \etal~\cite{smilkov2016embedding} developed \emph{Embedding Projector} as part of the TensorFlow framework~\cite{AbadiBarham2016}, enabling users to conveniently interact with embedding data and their local neighbors to quantitatively evaluate the embeddings.
Rauber \etal~\cite{rauber2016visualizing} employed t-SNE projections to visualize the hidden representations of deep neural networks across neural layers, to reveal how data instances of the same class progressively form clusters.  
Multiple visualization efforts aimed to disentangle the latent space of deep generative models by relating the latent dimensions with human-understandable visual concepts~\cite{wang2020scanviz,liu2019latent,liua2020latentvis}. 
The \textit{second} analytical task is to compare embeddings generated from different algorithms.
For example, \emph{embeddingVis}~\cite{li2018embeddingvis} focuses on graph embeddings and uses multiple juxtaposed t-SNE views to compare different embedding methods.
The same data instances are linked across all t-SNE views for explicit tracking.
Heimerl \etal~\cite{heimerl2020embcomp} proposed a set of metrics to measure the relationships (embedding correspondences) between two embeddings in a coordinated multi-view system called \emph{embComp}. 
Ghosh \etal~\cite{ghosh2020visexpres} introduced a toolkit -- \emph{VisExPreS} -- to disclose and compare the preserved global and local structures from the embeddings for novice data analysts. 

Our work covers both analytical tasks.
For semantic interpretation, we focus on the biases encoded in word embeddings and provide interactive applications that remove the biases through subspace transformations.
For embedding comparison, we allow any embedding to be analyzed before and after dynamic modification, and to compare how different debiasing techniques affect its  underlying geometry. 

Our work also intersects with visualization for ML fairness (e.g.,~\cite{KusnerLoftusRussell2017, CabreraEppersonHohman2019}).
It is specifically focusing on debias mitigation techniques in word embeddings. 
The \emph{What-If} tool~\cite{whatif} combines data exploration with counterfactual (what if) explanations~\cite{KusnerLoftusRussell2017} and fairness modifications. 
\emph{FairVis}~\cite{CabreraEppersonHohman2019} is used to audit pre-trained models for biases against known vulnerable groups in the context of a recidivism prediction system.
The \emph{DebIE}~\cite{friedrich2021debie} tool also illustrates bias in word representations but it limits itself to backend calculations of existing bias evaluations. 
A single interpretation of bias direction remains unknown, as well as extensions of debiasing techniques to contextual embeddings.

{\tool} integrates different aspects of debiasing in word representations with several debiasing techniques, encompassing bias subspace determination, generalizable~\cite{dev2020oscar} bias mitigation strategies, and bias measurement.
It helps highlight and compare the different combinations of subspace identification and bias mitigation strategies that work best for a given embedding and a particular type of bias.
{\tool} can also be utilized by users without previous knowledge or intuitions about potential biases to find issues within an incoming prediction model.

%% file: sec-debiasing.tex
\section{Debiasing Embedded Representations}
\label{sec:debiasing}

In this section, we review debiasing methods for embedded representations (e.g.,~\cite{MikolovSutskeverChen2013}), which  are becoming essential elements of many pipelines to process and understand texts and other types of complex data (e.g., merchant embeddings in~\autoref{sec:use-cases}).
The Euclidean nature of embedded representations makes them easy to integrate into existing analysis pipelines.
Furthermore, the similarity between the representations is encoded in a way that respects their complex interactions and often takes the nuance out of modeling and formalizing those notions.  

However, embedded representations also come with challenges.
Their \emph{distributed} nature means that the original features in texts  are no longer bound to specific dimensions.  Therefore their easy-to-interpret properties are now obfuscated.
A more troubling aspect is that these representations encode and potentially hide biases.
Caliskan \etal~\cite{CaliskanBrysonNarayanan2017} famously revealed that these representations encode well-trodden stereotypes, where male identifiers are more associated with careers and female identifiers are more associated with families.
In social studies of texts, as illustrated in~\autoref{fig:teaser}, adjectives such as ``rational'' and ``dependable'' are often used to describe male leaders while ``temperamental'' and ``excitable'' are often used to describe female leaders.
Such associations can be potentially harmful if these representations are used in tasks such as resume sorting, where potentially female candidates could be unwittingly given lower rankings because of this hidden gender bias.

As a result, over the last five years, there has been a rapid development of many debiasing mechanisms to pinpoint, quantify, and mitigate the biases within these representations.
While some have focused on correcting for these biases outside of the embeddings or in spite of them~\cite{zhao2018learning}, we focus on approaches that directly analyze and mitigate issues in these embedded representations.
First, these representations are widespread and so the techniques are highly transferable.
Second, the techniques are simple and efficient, and can be applied dynamically depending on the tasks, including for contextual embeddings~\cite{dev2019measuring}.

In developing {\tool}, we identify that these mechanisms to analyze and mitigate biases in embedded representations can be decoupled into a two-step process.
The first is to identify a concept subspace among the vectorized representations that capture the direction of bias (e.g., the concept of gender or nationality).
The second is to use this subspace to transform the representations in a simple and controlled way.

These debiasing approaches may require the knowledge of one or two concept subspaces, or additional word lists for evaluation.
For the most part, their comparative advantages, disadvantages, and relations are not often discussed, leading to rediscoveries of ideas and potential confusions among the users.
Using {\tool} as an educational medium, we first review concept subspace identification methods, and then describe how these subspaces are used in removing or disassociating these concepts from the embedded representations.

\subsection{Methods of Subspace Identification}
In embedded representations, the specific dimensions occupied by features are unknown. In this section, we discuss four methods in the literature used to determine the subspace that is the span of a specific concept (e.g., gender).
Some of these methods (PCA and PCA-paired) naturally generalize to identify multiple directions, but it is quite rare to use more than one direction to represent a concept.
To keep subspace identication modular and simple, VERB currently  identifies one-dimensional subspaces, as described below.

\begin{figure}[t]
\centering
\includegraphics[width=0.98\columnwidth]{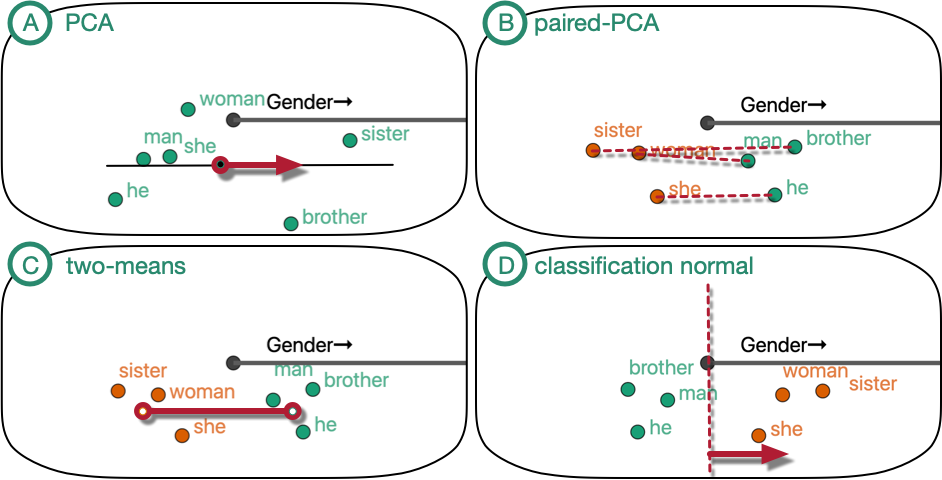}    
\vspace{-3mm}
\caption{Concrete examples of subspace identification methods using {\tool}. From left to right: PCA, paired-PCA, 2-means, and classification normal. (B-D): green points represent male gendered words, orange points represent female gendered words, and the black line segment that starts from the origin represents the gender subspace direction.}
\label{fig:subspaces-verb}
\vspace{-2mm}
\end{figure}

\para{PCA.}
The most general and simple approach to determine a subspace is referred to as the PCA method.
It requires one set of word vectors, referred to as the \emph{seed words}, of which it computes the top principal component -- which is the best one-dimensional subspace that minimizes the sum of squared distances from all word vectors.
This resulting unit vector represents the subspace direction.
Using {\tool}, we illustrate the PCA method in~\autoref{fig:subspaces-verb}(A) using a set of gendered seed words:``man, woman, brother, sister, he, she''.
The arrow above the black line segment points towards the direction of the gender subspace obtained via PCA.

\para{Paired-PCA.}
Another variant based on PCA was proposed by Bolukbasi \etal~\cite{BolukbasiChangZou2016}.
It requires a list of paired words as the seeds, each pair has one word vector from different groups.
For example, for the gender concept shown in ~\autoref{fig:subspaces-verb}(B), we use ``man-woman, he-she, brother-sister'' as seeds for subspace identification.
Paired-PCA method then reports the concept subspace as the first principle component of the difference vectors between each paired vectors.
A subtle note: because these vectors are the result of differences, we do not need to ``center'' them (remove their mean) first as when PCA is used on word vectors.

\para{2-Means method.}
The 2-means method~\cite{DevPhillips2019}, for any two sets of words as seed sets, returns the normalized difference vector of their respective averages.
So for groups of words $F=\{f_i\}$ and $M=\{m_i\}$, it computes $f = \frac{1}{|F|}\sum_{i} f_i$ and $m = \frac{1}{|M|}\sum_{i} m_i$ as the mean of each set respectively.   
Then the direction is calculated as
$
v = \frac{f - m}{\|f - m\|}.  
$
This method has the advantage that it does not require paired words or an equal number of words in the two seed sets.
We give an example of applying the 2-means method to two sets of seed words in~\autoref{fig:subspaces-verb}(C), where $F=\{``woman", ``sister", ``she"\}$ and $M=\{``man", ``brother", ``he"\}$.
The computed gender direction $v$ originates from the origin in the visualization.

\para{Classification normal.}
For two groups of seed words that can be classified using a linear SVM, the direction perpendicular to the classification boundary represents the direction of difference between the two  sets.
Again, this only requires two sets $F$ and $M$, but they do not need to be paired or of equal size.
This is illustrated in~\autoref{fig:subspaces-verb}(D), where the dotted line represents the classification boundary between $F=\{``woman", ``sister", ``she"\}$ and $M=\{``man", ``brother", ``he"\}$.
The black segment emanating from the origin again indicates the gender direction.
Ravfogel \etal~\cite{ravfogel2020null} used this direction iteratively to remove bias in word vectors by projections.

\subsection{Bias Mitigation Methods}
There are several methods to modify the embedding structure in ways that mitigate the encoded bias.
While there are more complicated optimization-based ones designed for specific tasks in gender bias in text~\cite{zhao2018learning}, we describe a subset of four debiasing methods that are quite simple to actuate (although nuances of them may be confusing), and rely specifically on the concept subspaces identified earlier.
Again, {\tool} serves as the perfect visual medium to explain these debiasing methods.
For the descriptions below, a point in the space of high-dimensional embedded representations is denoted as $x \in \Rspace^d$ (e.g. for $d = 50$ or $d=300$).
A concept subspace is labeled $v$ and is restricted to be a unit vector in $\Rspace^d$. 

\para{Linear Projection (LP).}
The simplest approach~\cite{DevPhillips2019} removes the component of concept subspace for each data point $x$.
This can be applied individually to each data point $x$, where the component along $v$ is $\langle v,x \rangle v$, where $\langle v, x \rangle$ is the Euclidean dot product.
The LP method then removes the component along $v$ for every point $x \in \Rspace^d$ as
$
x' = x - \langle v,x \rangle v.  
$

\begin{figure}[!h]
\centering
\includegraphics[width=0.98\columnwidth]{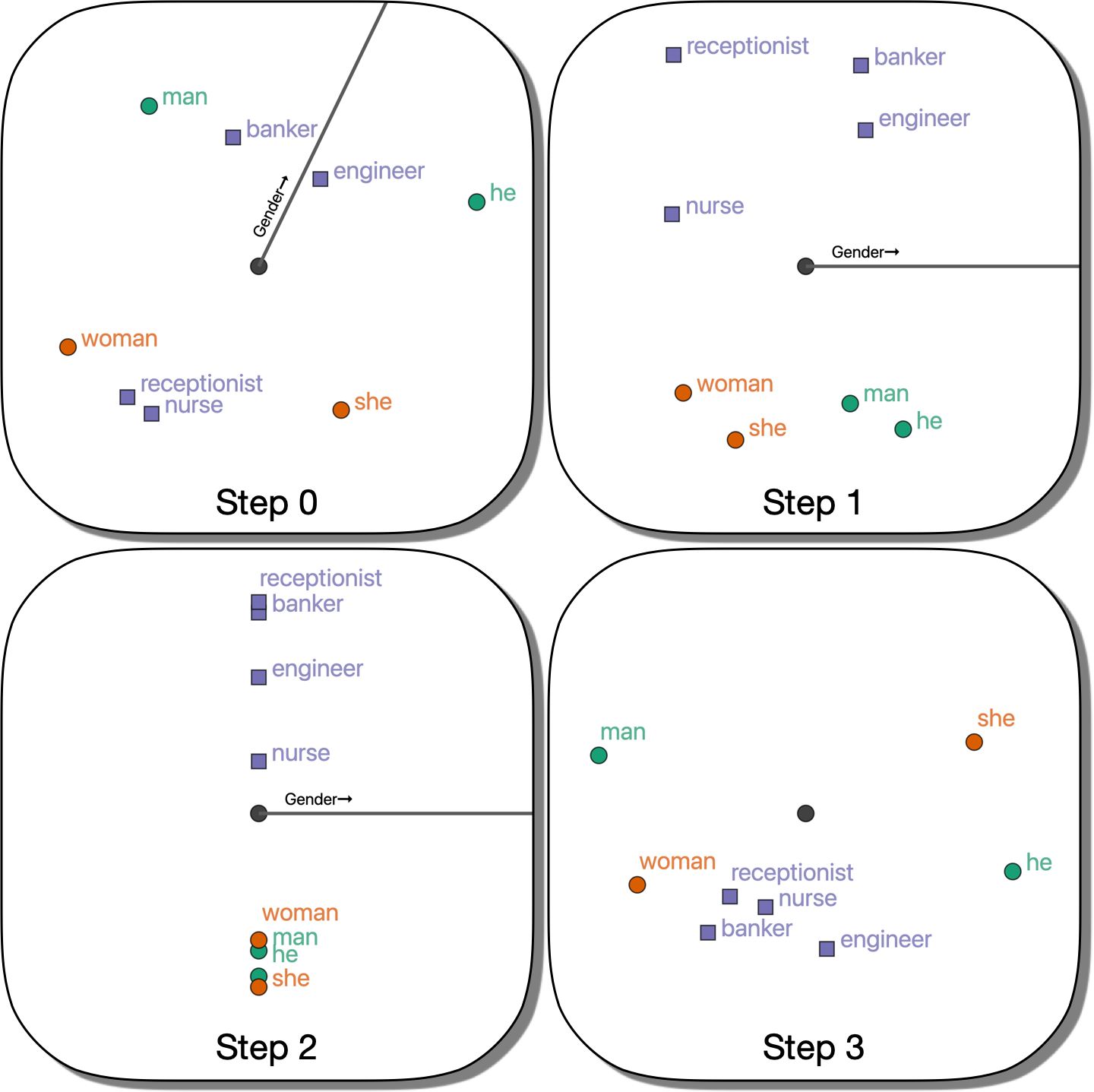}
\vspace{-3mm}
\caption{A simple example using two-means to identify a gender subspace and Linear Projection to mitigate the gender bias in word embeddings.}
\label{fig:example-lp}
\end{figure}

Using {\tool}, we give a simple example, by applying two-means and LP debiasing in mitigating the gender bias in occupational words.
The two seed sets are $M$=\{``man", ``he"\} and $F$=\{``woman", ``she"\}. The evaluation set is $E$ = \{``receptionist", ``nurse", ``scientist", ``mathematician"\}.
As illustrated in~\autoref{fig:example-lp}, {\tool} decomposes the LP method into an interpretable sequence of transformations.
In step 0, both seed sets and evaluation set are viewed using a perspective from PCA, where the gender direction is identified using two-means.
In step 1, the viewing perspective/angle is reoriented so the gender direction is aligned with the x-axis, where we see clearly that ``receptionist" and ``nurse" are shown to be closer towards the female direction while ``banker" and ``engineer" are closer towards the male direction. 
In step 2, for every word in the embedding, LP removes its component along the gender direction in $\Rspace^d$, where all words are shown to be aligned along the horizontal axis.
The underlying data is modified in this step.
In step 3, the transformed (debiased) points are reoriented again using the perspective from PCA, where there is no clear gender association among the occupational words.
This is different from the original view since the data was modified in step 2.

\para{Hard Debiasing (HD).}
An earlier approach (the first one proposed) by Bolukbasi \etal~\cite{BolukbasiChangZou2016}, known as Hard Debiasing, uses a similar mechanism, and is designed specifically for gender bias.
It also requires an additional wordlist called the \emph{equalize set}, which are used to preserve some of the information about that concept.
We summarize this mechanism next.
The words that are used to define $v$ are considered definitionally gendered and not modified.
The exception is another provided set of pairs of words (e.g., ``boy-girl", ``man-woman", ``dad-mom", ``brother-sister").  
These word pairs are \emph{equalized}; that is, they are first projected as in Linear Projection, but then each pair is extended along the direction $v$, so the words are equally far apart as they were before the operation.
The remaining words are then projected as in Linear Projection.

\begin{figure}[!h]
\centering
\includegraphics[width=0.98\columnwidth]{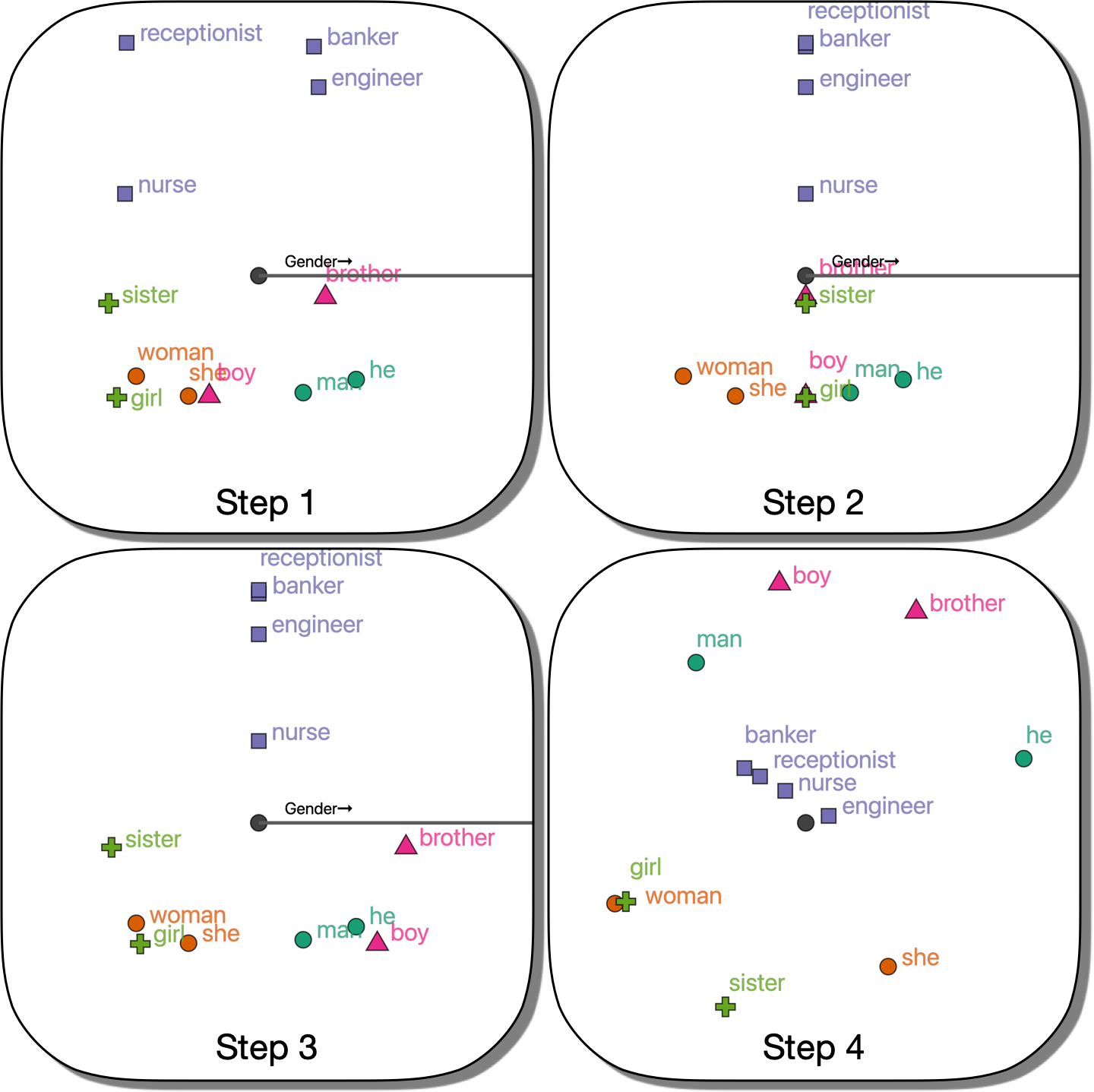}
\vspace{-3mm}
\caption{An example using two-means and HD to mitigate the gender bias.}
\label{fig:example-hd}
\end{figure}

In our example with {\tool}, we again use $M$=\{``he, man"\} and $F$ =\{``she, woman"\} and two-means to define a gender direction $v$, $Q$= \{``boy-girl",``sister-brother"\} as the equalize set, and $E$= \{``engineer'', ``lawyer'', ``receptionist'', ``nurse'' \} again as the evaluation set.  
This is illustrated in~\autoref{fig:example-hd}. 
Step 1 is obtained after  a reorientation of the gender direction along the x-axis. 
Step 2 is removing the component of each point along the gender direction with the exception of $M$ and $F$ (``she, woman'' and ``he, man").
Step 3 tries to preserve some information regarding gender using the equalize set $E$ thus extending the words in $Q$ (``brother", ``sister", ``boy", ``girl") along the gender direction so they become equally far apart. 
Step 4 reorients the modified words using PCA from a viewing perspective with the most variance.

Bolukasi \etal~\cite{BolukbasiChangZou2016} described other methods, and later works by Wang \etal~\cite{wang2020double}  also provided slight variants, or rediscovered these approaches.
One concern about Hard Debiasing is that it may leave residual bias~\cite{gonen2019lipstick}.
The authors of that critique helped develop the next approach as an alternative.

\para{Iterative Nullspace Projection (INLP).}
INLP~\cite{ravfogel2020null} starts with a pair of large word lists (e.g., sets of male and female words). 
It suggests to select the top $0.5\%$ of the extreme words along either directions of the he-she vector, denoted as sets $M$ and $F$ respectively. 
It then builds a linear classifier that best separates $M$ and $F$, and linearly projects all words along the classifier normal (denoted as $v_1$). 
However, a classifier with accuracy better than random may still be built on $M$ and $F$ after the projection, let $v_2$ denote the classifier normal. 
INLP then applies linear projection to all words again along $v_2$.  
This continues for some large number of iterations (their code uses 35 iterations).  
Afterwards, the words which may encode bias, even by association (the sets $M$ and $F$), cannot be linearly separated with accuracy better than random chance.  

\begin{figure}[!h]
\centering
\includegraphics[width=0.98\columnwidth]{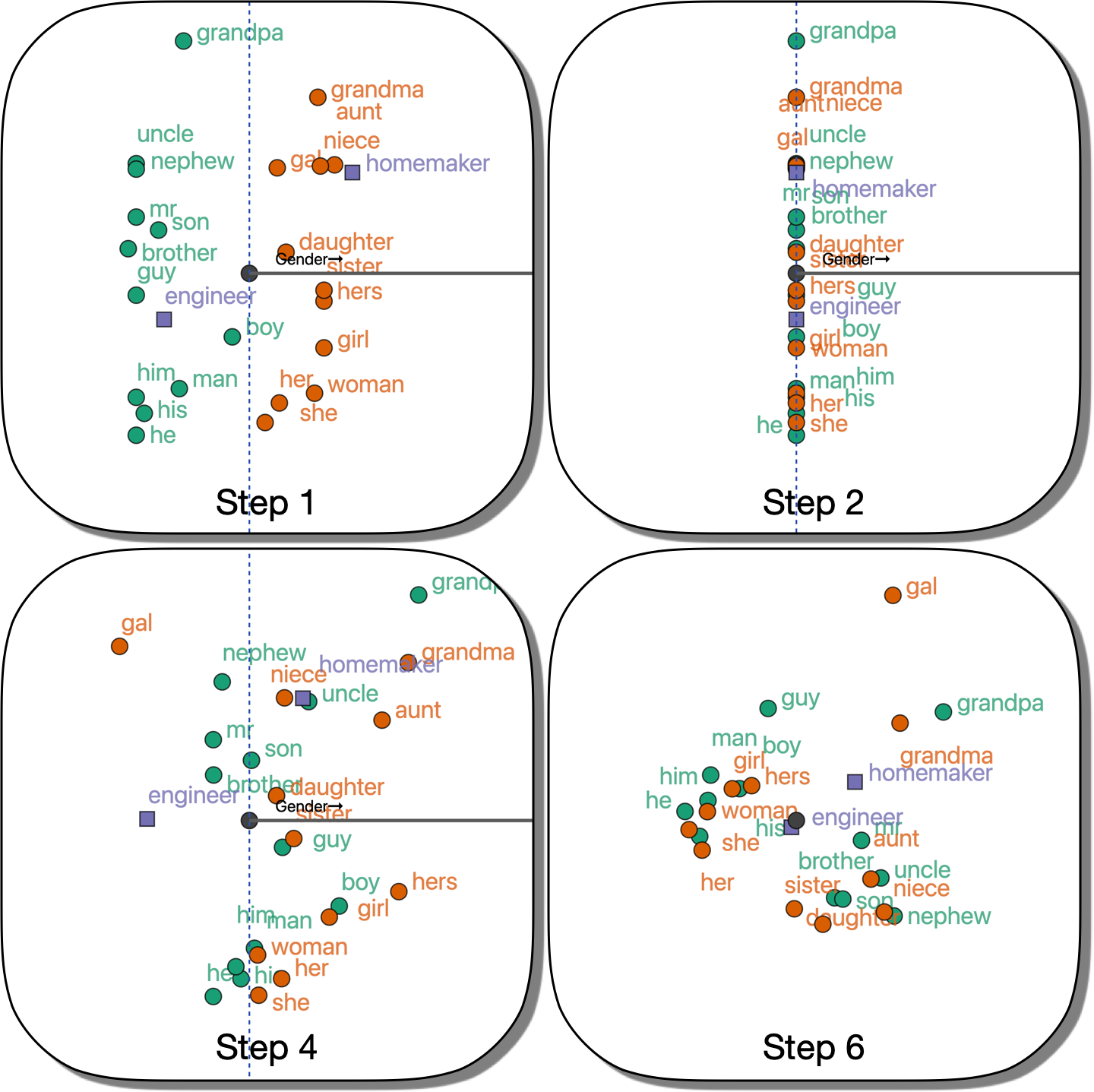}
\vspace{-3mm}
\caption{An example using classifier normal and INLP to mitigate the gender bias over two rounds.}
\label{fig:example-inlp}
\end{figure}

An example run of INLP using {\tool} is shown in~\autoref{fig:example-inlp} using two sets of definitionally gendered words $M$ = \{``man, he, him, his, guy, boy, grandpa, uncle, brother, son, nephew, mr"\} and $F$ = \{``woman, she, her, hers, gal, girl, grandma, aunt, sister, daughter, niece"\}.
A perfect separator/classifier can be found initially (shown in Step 1), and then linear projection along the classifier normal is shown in Step 2.
The next classifier normal (shown in Step 4) is not a perfect separator.
Yet after its next application, and a PCA reorientation as shown in Step 6, no sufficiently good classifier can be found, and the procedure stops.

\para{Orthogonal Subspace Correction and Rectification (OSCaR).}
A critique of the above techniques, especially INLP, is that they are destroying information that we might want to preserve. 
For example,  we may want to know that ``grandpa'' is referring to a male grandparent.  
The OSCaR approach~\cite{dev2020oscar} seeks a more controlled approach. 
It requires two specific concept subspaces, for instance, one  representing gender $v_1$ and another representing occupations $v_2$.  
OSCaR does not project out the gender subspace, but rather attempts to disassociate them by making those subspaces \emph{orthogonal}.  
In addition to orthogonalizing those subspaces, which can be done by rotating $v_2$ to $v_2'$ so $\langle v_1, v_2' \rangle = 0$, it also rotates all other data points by a lesser amount.  
Points close to $v_1$ do not rotate much, while points close to $v_2$ rotate about as much as $v_2$.  
While OSCaR does not remove any possible way to find any association between data aligned with either of these subspaces, it does make the concepts as a whole orthogonal. 
In the bias evaluation approaches described in~\autoref{sec:bias-eval}, OSCaR is demonstrated to reduce bias in an amount similar to  other debiasing approaches.  
Moreover, it retains the information along each of the original subspaces $v_1$ and $v_2$.  

\begin{figure}[!h]
\centering
\includegraphics[width=0.98\columnwidth]{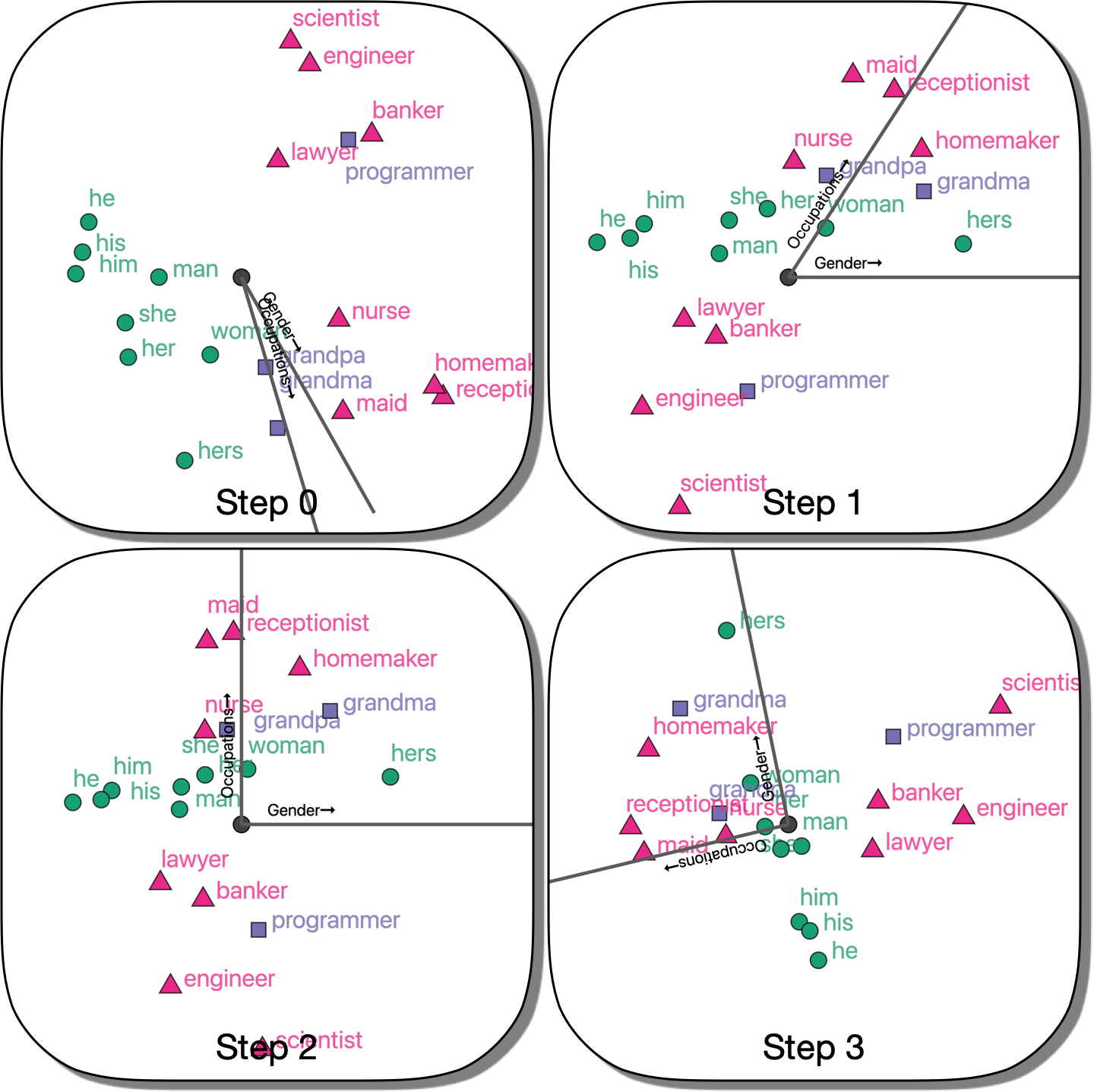}
\vspace{-3mm}
\caption{An example using PCA and OSCaR to rectify gender bias in relation to occupations.}
\label{fig:example-oscar}
\end{figure}

With {\tool}, \autoref{fig:example-oscar} shows the four steps of OSCaR.  
The first subspace $v_1$ representing gender is defined with words ``he", ``his", ``him", ``she", ``her", ``hers", ``man", and ``woman."  
The second subspace $v_2$ representing occupations is defined with words ``engineer", ``scientist", ``lawyer", ``banker", ``nurse", ``homemaker", ``maid", and ``receptionist."  In the PCA view (Step 0), one can observe that the two subspaces are correlated, and the typical gender stereotypes of the occupation is present in the word representation, e.g., ``maid" towards the female and ``engineer" towards the male direction.  
The reoriented view in Step 1 aligns the Gender direction ($v_1$) along the x-axis. It shows the span of $v_1$ and $v_2$, which is the 2-dimensional subspace where OSCaR modifies the data.
It is also the subspace with the largest angle between these two subspaces.
In Step 2, the data is modified so that the gender and occupation subspaces become orthogonal.
The Evaluation set words ``grandma", ``grandpa", and ``programmer" (along with all other words), can be seen to move along with these words.
Note how ``programmer" is still near the other technical-oriented careers, and how ``grandpa-grandma" retains the inherently male-female relationship.
Finally, in Step 3, another PCA view is shown on the modified data, and now the subspaces can be seen to retain the orthogonal nature, and the gender connotation in the occupations has been rectified, so there is no apparent stereotypical correlation.  

\subsection{Bias Evaluation Methods}
\label{sec:bias-eval}

There are several intrinsic~\cite{CaliskanBrysonNarayanan2017,DevPhillips2019} and extrinsic~\cite{zhao2018learning,dev2020measuring} measures to determine how much bias is contained by word embeddings.
When bias is removed~\cite{BolukbasiChangZou2016,dev2020measuring,dev2020oscar}, these measures help determine how effective the bias removal has been.
In general, it may not be possible to completely remove bias in these measures due to the nature of the measurement or its influence from other data and training choices.
We next describe some common and representative bias measurement methods.

\subsubsection{WEAT}
\label{weat}
The Word Embedding Association Test (WEAT)~\cite{CaliskanBrysonNarayanan2017} is an analogue to Implicit Association Test (IAT) from psychology.
It checks for human like bias associated with words in word embeddings.
For example, it found career oriented words (e.g., ``executive", ``career") are more associated with statistically male names (e.g., ``Tom", ``Peter") and male gendered words (e.g., ``man", ``boy"); while family oriented words (e.g., ``family", ``home") are more associated with statistically female names (e.g., ``Mary", ``Kate") and female gendered words (e.g., ``women", ``girl"). 

WEAT considers four sets of words:  two target word sets $X$ and $Y$ (e.g., representing male and female genders) and two sets of attribute words $A$ and $B$ (e.g., representing stereotypical male or female professions). 
First, for each target word $w \in X \cup Y$, it computes how much the word is associated with set $A$, and not associated with set $B$ as
\[
s(w,A,B) = \frac{1}{|A|}\sum_{a \in A} \cos(a,w) - \frac{1}{|B|}\sum_{b \in B} \cos(b,w),
\]
where $\cos(a,w)$ is the cosine similarity between vector $a$ and $w$.    
Then it averages this across all $w \in X$, minus the average of all $w \in Y$ as
\[
s(X,Y,A,B) = \frac{1}{|X|}\sum_{x\in X} s(x,A,B) - \frac{1}{|Y|}\sum_{y\in Y} s(y,A,B).  
\]
Finally, the WEAT test statistic is $s(X,Y,A,B)$ normalized by the standard deviation of $s(w,A,B)$ for all $w \in X \cup Y$, so typical values should not be too far from $[-1,1]$, and a value closer to $0$ indicates less implicit (and biased) association.  

As the single most common bias quantification, {\tool} allows users to compute WEAT before and after debiasing.  The default word sets (which can be modified) in {\tool} use the following~\cite{CaliskanBrysonNarayanan2017, DevPhillips2019}:  
\begin{itemize}\denselist
\item Male words as X = \{male, man, boy, brother, he, him, his, son\} 
\item Female words as Y = \{female, woman, girl, sister, she, her, hers, daughter\}
\item Stereotypically male occupations A = 
\{doctor, engineer, lawyer, mathematician, banker\}
\item Stereotypically female occupations B = 
\{homemaker, receptionist, dancer, maid, nurse\}
\end{itemize}

\subsubsection{Embedding Coherence Test}
The Embedding Coherence Test (ECT)~\cite{DevPhillips2019} measures if groups of words have stereotypical associations.
Instead of evaluating the exact word similarities (e.g., male and female words with occupation words), it first aggregates the male and female words into their means, described as $m = \frac{1}{|X|}\sum_{x \in X}x$ and $f= \frac{1}{|Y|}\sum_{y \in Y}y$.  Then it evaluates if the order of similarity from $m$ and from $f$ to a different set $A \cup B$, such as occupation words (``doctor", ``nurse", etc.).
Then it sorts the values $\cos(m,w)$ for each $w \in A \cup B$ and the values $\cos(f,w)$.
The similarity of these sorted lists is measured with the Spearman Coefficient, which ranges between 1 (when the ordering is exactly the same, so the least bias) and -1 (where the ordering is exactly opposite, thus the most bias).
So larger values of ECT indicates less bias.  

\subsubsection{NLI Based Tests}
\label{sec: NLI test}
Since word representations are used downstream in different tasks and applications in NLP, it is important to measure the effect biased associations have on the decisions made in these tasks.
An example is natural language inference (NLI). Dev \etal~\cite{dev2019measuring} used NLI to provide a clear signal on the encoded bias.
The task is, given a pair of sentences to predict if the second one is entailed, is contradicted, or is neutral to the first sentence. 
The sentence pairs constructed as all neutral and any deviation from a neutral prediction is bias.
These are constructed from simple template sentences where the verb, object, and subject are chosen from word lists, and in total over a million sentences are considered.
For each one, the subject (e.g., an occupation like ``doctor'') is paired with another sentence where the subject is replaced by either ``man'' or by ``woman''.
If the occupation has no gender bias, it will result in a neutral inference prediction, but if bias is encoded, it will result in higher probability of entailment or contradiction.
The higher the percentage of sentences predicted as neutral, the better the prediction (i.e., the lower the amount of bias).

%% file: sec-tool.tex
\section{The {\tool} User Interface}
\label{sec:tool}

We present {\tool}, an interactive system for visualizing and interpreting bias mitigation techniques for word representations.

With {\tool}, users can explore and interpret four types of debiasing techniques through three coordinated views. 
The \textbf{Embedding View} (\autoref{fig:teaser}A) highlights a subset of word vectors using dimensionality reduction and visualizes the transformations of their embeddings across various debiasing techniques. It decomposes a chosen technique into a sequence of interpretable operations and visualizes their associated transformations via a step-by-step animation. 
It also provides additional capabilities to interact with individual word vectors. In particular, users can select a word in the Embedding View and {\tool} will display its nearest neighbors in the high-dimensional embedding space, before and after debiasing.    
The \textbf{Control Panel} (\autoref{fig:teaser}B) enables users to configure each debiasing technique by specifying the algorithm (LP, HD, INLP, or OSCaR), subspace technique (PCA, paired-PCA, 2-means, or classification normal), concept labels (e.g., gender, occupation), seed sets (for defining concept directions), evaluation set, and equalize set, etc. 
It also provides controls to navigate through the steps of the chosen  technique and to toggle various aspects of the visualization such as data labels, subspace direction, and evaluation points. 
Users can also choose from a list of predefined examples (detailed in~\autoref{sec:use-cases}).
The \textbf{Explanation Panel} (\autoref{fig:teaser}C) gives a step-by-step description of the transformation. 
Finally, {\tool} enables users to download the modified word embedding after applying a particular debiasing technique. 
Thus it not only provides an educational guide for understanding a debiasing technique, but also allows users to apply and visually verify these modifications  before moving to downstream analysis.

In the example shown in \autoref{fig:teaser}, {\tool} reveals association in the word embedding with the \emph{gender} concept that contributes to its gender bias.
Specifically, before debiasing (\autoref{fig:teaser}A), along the gender direction (a black line segment that starts from the origin), the adjectives ``strong'', ``important'', ``arrogant", ``rational'' are more closely associated with the male words ``he'' and ``him", while the words ``temperamental'', ``gossip'', ``excitable'', and ``beautiful'' are more closely associated with the female words ``she'' and ``her''.  
{\tool} then animates the step-by-step transformation of the word embedding using LP for debiasing and two-means for subspace identification.  
After debiasing (\autoref{fig:teaser}D), there is no longer a gender direction, and the above adjectives do not provide a clear gender bias.

\para{Transforming embedding views.}
A central functionality of {\tool} is that it allows users to experiment with and visualize the effects associated with a chosen debiasing mechanism. 
Specifically, the tool updates the Embedding View when an algorithm modifies the underlying representation step-by-step. 
As we demonstrate in~\autoref{sec:use-cases}, {\tool} can be applied to not only word vector embeddings that arise from NLP, but also other embedded representations.  
While the data is represented as 50- or often higher-dimensional vectors, {\tool} provides views of the data objects in a 2-dimensional interface as points. 
While our default embedding has 100K points (others could have much more), we do not attempt to visualize all of these points.
Instead, the Control Panel allows users to select a representative subset (i.e., the evaluation set) to visualize.  

After choosing a debiasing mechanism and a subspace identification technique in the Control Panel, each debiasing process always starts with a 2-dimensional perspective, determined by the best 2-dimensional subspace as determined by PCA on the user-provided data points.  
The origin is always shown in the center of the Embedding View. 
 Since the data points are often interpreted as vectors, the cosine metric is the most commonly used metric (which measures angles with respect to the origin as a base point).  
This initial PCA view (marked as ``step 0'' in the Embedding View) itself is not especially meaningful, but useful as an alternative starting point, and helps highlight the meaningfulness of the other views.  

In ``step 1'' of the transformation, {\tool} changes its viewing perspective of the initial embedded view.  
In particular, the concept subspace determined by $v$ is always rotated so it is aligned with the $x$-axis.
The $y$-axis is chosen as the highest variance direction among the remaining points (via PCA).  This choice of $x$-axis is essential for two reasons.
First, the left-right direction provides a faithful account of how far each representative point is along this concept subspace.
Two related terms (e.g., ``temperamental'', ``rational'') can be compared, and their relation along this concept subspace (e.g., gender) is not distorted, which may be the case if that subspace was not parallel with the viewing plane.
Second, when a projection operation (internal to three of the debiasing mechanisms) is applied that effectively removes the component along this concept subspace, then one can clearly see the representative data points moving onto a lower-dimensional subspace combined into and represented by the $y$-axis.

The OSCaR mechanism requires the definition of two subspaces $v_1$ and $v_2$.
In this case, the initial embedding view is the span of these two subspaces.
In OSCaR, all of the operations happen only in this subspace, while all components outside this 2-dimensional subspace (e.g., coming out of, or into the screen) are not modified.
As with all techniques, this allows users to explicitly see the action happening without any visual side-effects which obscure these operations.

For the INLP operation, which iteratively applies linear projection after finding the new best classsifier, {\tool} shows each of these steps.
After each new normal direction is found for a subspace, it updates the viewing perspective to make that subspace along the $x$-axis as before, so that residual concept can be viewed, and its projection can be dynamically visualized.  

Finally, with the new (modified) embedded representations, {\tool} then provides a final view of the best 2-dimensional PCA perspective of the data.
This is important to show the final and best possible view of the representations, especially when a project mechanism is used, and otherwise all of the data may have been compressed into a single 1-dimensional subspace (the $y$-axis) within that perspective.  

\para{Implementation.}
The front end of {\tool} is implemented using the HTML/CSS/Javascript stack and D3.js.
We use an automatic label placement algorithm \cite{Wang2013} which uses simulated annealing to minimize overlaps between text labels in the Embedding View.
Its back end is developed using Python and Flask.
{\tool} comes (by default) with a 50-dimensional GloVe embeddings of the 100K most frequent words taken from the Wikipedia 2014 + Gigaword corpus~\cite{PenningtonSocherManning2014}.
It also provides a larger GLoVe embedding (300-dimensional with the 100K most frequent words) from the Common Crawl corpus (https://commoncrawl.org/).

%% file: sec-use-cases.tex
\section{Use Cases}
\label{sec:use-cases}

We demonstrate the efficacy of {\tool} on several examples.  
First, we will show how it can be used to quickly and easily identify new forms of bias in word vector embeddings.  
Second, we will demonstrate the power of {\tool} in teaching and contrasting methods to identify concept subspaces and use them to attenuate bias in word vector embeddings.  
Third, we will highlight how {\tool} identifies concept subspaces as a critical yet under-explored element of debiasing. 
Inspired by this, we will show how to optimize subspace identification, leading to an improved iterative method which quantitatively improves the debiasing results.  
Finally, we will showcase {\tool}'s generality, by exploring a different type of embedded representation, one that captures merchant embeddings associated with a large payment company.

\subsection{Using {\tool} to Identify New Types of Biases}

{\tool} allows users to load any word vector embeddings or embedded representations (e.g., GloVe embeddings and merchant embeddings). 
Users can select any subset of words or identifiers (e.g., evaluation set) to quickly illustrate potential correlations.  
We demonstrate how {\tool} can be used to identify new types of biases such as the \emph{royalty bias}.  

\begin{figure}[t]
\includegraphics[width=0.49\linewidth]{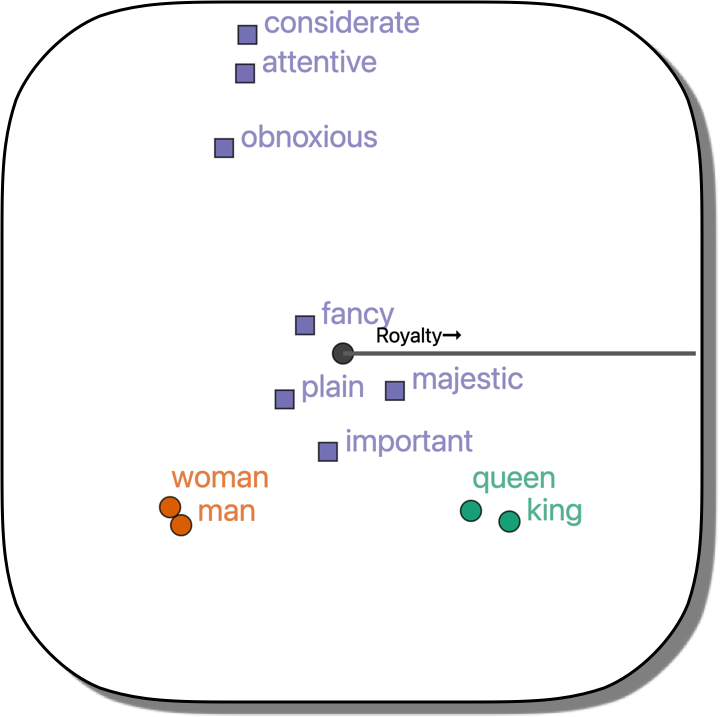}
\includegraphics[width=0.49\linewidth]{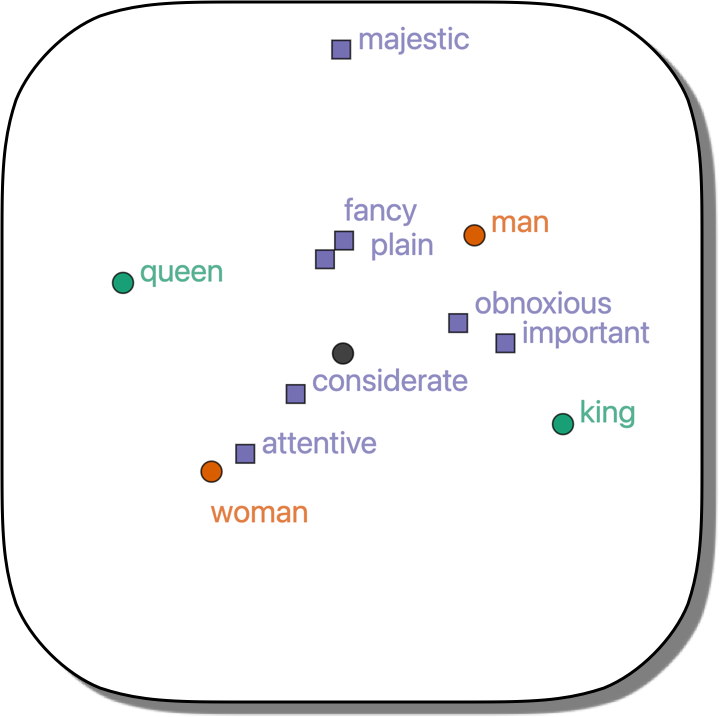}
\vspace{-2mm}
\caption{Left: royalty bias as observed in adjectives, e.g., ``majestic'' for ``queen, king'' vs. ``obnoxious'' for ``women, man''. Right: it shows residual gender associations after removing the royalty subspace using LP.
}
\label{fig:royalty}
\end{figure}

As an example, using {\tool}, one can easily observe that word embeddings capture a clear royalty subspace and resulting bias.
For instance, using two-means with seed words ``king, queen" (for royal words) and ``man, woman'' (for common words), a clear royalty subspace becomes quickly apparent.
If users also visualize some adjective words as the evaluation set, ``obnoxious, considerate, plain, fancy, attentive, important, majestic'', as in~\autoref{fig:royalty}, they can see a potential bias arising in the captured connotation.  
Words ``obnoxious, attentive, plain, considerate'' are more associated with the common direction, while ``fancy, important, majestic'' more in the royal direction.  

Note that the y-axis is chosen to show the most variance, and that variation along that direction is not correlated with royalty.
Whereas, the x-axis is selected to reflect the learned royalty component, and the further left along this coordinate the more common, and the more right the more royal association.

Moreover, after removing the royalty concept subspace with linear projection (LP), then through {\tool}, users can observe that the gender concept remains, as shown in~\autoref{fig:royalty} (Right).
After another PCA-based re-orientation, the words ``man'' and  ``king'' are to the right, and words ``woman" and ``queen" to the left.
Also, there is still residual gender associations after debiasing.
Stereotypical male, chauvinistic traits ``important, obnoxious" are more on the male side while the stereotypical female subservient traits ``considerate, attentive" are more associated with the female side.

\subsection{Using {\tool} to Explain Debiasing Methods}

{\tool} was used as part of an AAAI 2021 Tutorial titled ``A Visual Tour of Bias Mitigation Techniques for Word Representations''.
{\tool} is designed to target NLP practitioners who are designing decision-making systems with word embeddings, and also researchers working with fairness and ethics of ML systems in NLP.
It also serves as a visual medium for education, which helps NLP novices to understand and mitigate biases in word embeddings. 

To explain diabsing methods to these targeted audience, the description of the debasing methods alone makes it hard for an interested user to decide which one to use, how to use them, and what the limitations are.  
During this tutorial, participants could easily download {\tool}, and immediately start interacting with the pre-loaded word vectors embeddings, or creating new examples.  
They can not only clearly observe the biases and structures presented in these embedded representations, but also compare and contrast their effectiveness and side effects.   

While {\tool} provides utility to run a variety of identification of concept subspaces, and then uses them within debiasing approaches, we highlight a few examples where it is particularly effective in  distinguishing variation in the debiasing methods and improving users' understanding of them. 

\para{Comparing Hard Debiasing vs. Linear Projection.}
Bolukbasi \etal~\cite{BolukbasiChangZou2016} introduced the idea of using the Linear Projection (LP) step towards debiasing, but wrapped it in a more complex Hard Debiasing (HD) mechanism in an attempt to preserve the structure among definitionally gendered words.  
This mechanism requires extra word lists and includes a set of paired words, which are equalized.
To demonstrate the difference, we use {\tool} to run HD and LP in \autoref{fig:example-hd} and \autoref{fig:example-lp}, respectively, using the same seed sets to define the subspace, and evaluation set.
One can easily observe that HD requires an extra equalize set, and as a result an extra step in the process to equalize those pairs.
Hence, it also requires the concept must be the result of some binary notion, thus disallowing concepts like nationality.
On the other hand, LP simply projects all words to a subspace that is one dimension lower, including the seed set.
While these methods are distinct, it may not be clear all the ways they differ without {\tool}.

\para{Understanding OSCaR.}
OSCaR~\cite{dev2020oscar} is a new approach to debiasing, and does not rely on projection to remove a subspace.
Instead, its operation focuses on a graded rotation (where a different rotation matrix is applied to each point) that, while subdifferentiable, requires a complicated case statement to define precisely.
With {\tool}, users are able to, for the first time, dynamically visualize this process under a number of situations.
\autoref{fig:example-oscar} show snapshots of the process on an example.
In particular in Step 1, users are presented the specific perspective needed to understand the graded rotation, and then it is animated between Steps 1 and 2.
Furthermore, users can see how afterwards, both the concepts remain in tact, but they have had their correlation removed.

One observation that quickly became apparent with {\tool}, but not before, is that OSCaR works more closely to what one's intuition might be (of orthogonalizing subspaces) when the subspaces are defined using PCA.  
While two-means may do a better job of explicitly capturing the concept subspace for the relationship between two sets (e.g., definitionally male and female words), they do not as explicitly capture a single subspace for the concept as does PCA.  
Visually, using PCA with OSCaR allows users to easily see the subspace for each concept, where the one for gender can be seen as much less noisy than one for occupations, and how they are orthogonal after the operation.

\begin{figure}
\includegraphics[width=0.49\linewidth]{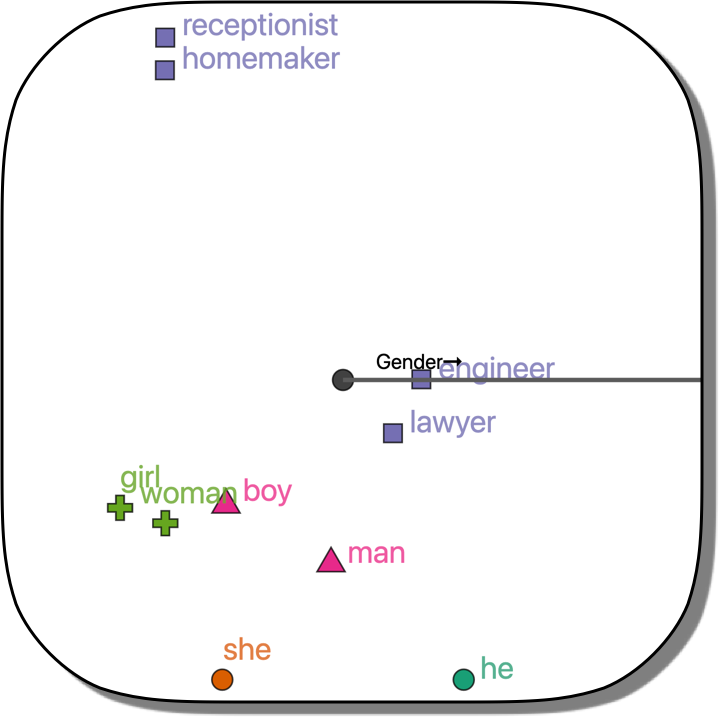}
\includegraphics[width=0.49\linewidth]{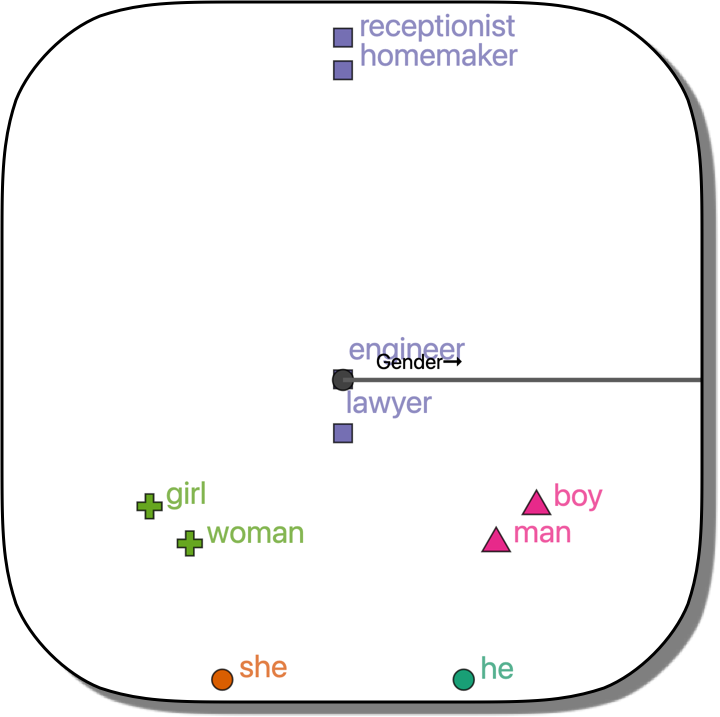}
\vspace{-2mm}
\caption{{\tool} visualizes residual bias after applying Hard Debiasing.}
\label{fig:residual}
\end{figure}

\para{Residual Bias.}
A well-known critique of Hard Debaising~\cite{gonen2019lipstick} is that it leaves residual bias in the embedding, even after the debiasing operation.
While this is illustrated mostly quantitatively or abstractly in~\cite{gonen2019lipstick}, with {\tool}, users can easily see the potentially concerning effect.
For example, when trying to remove gender bias associated with occupations, HD projects occupation words off of the gender-defining subspace.  However, for instance, as seen in~\autoref{fig:residual}, the traditional and stereotypical female occupations (e.g., ``receptionist'', ``homemaker'') are still very close to one another, as are stereotypical male professions (e.g., ``lawyer'', ``engineer'').
This illustrates the concern since if one knows a homemaker is traditionally female and an engineer is not, then one may infer that so is receptionist, and a lawyer is also not.

\subsection{{\tool} Inspired Interactive Concept Identification}
In building {\tool}, we realize that it provides a new abstraction for understanding existing and creating new debiasing methods. 
In particular, {\tool} uses a two-step process, subspace identification and embedding transformation, for a debiasing process. 
Instead of transforming a word vector embedding with a pre-defined concept subspace, we study how to (a) determine an ``optimal'' concept subspace and (b) apply transformation {\wrt} such a subspace.

As reviewed in~\autoref{sec:debiasing}, there are several distinct methods to identify concept subspaces (PCA, paired-PCA, two-means, classifier normal).
Although these methods are often associated with specific debiasing methods, they are mostly interchangeable.
Despite of their distinctiveness, the choice of subspace identification method is often ignored or neglected.
{\tool} allows a users to easily experiment with these methods, explore their differences, their sensitivity to seed words, and improve the overall effectiveness.
Then, going beyond existing mechanisms, {\tool} provides a novel, optimized concept identification method, improving upon prior methods in quantifiable bias mitigation.

\subsubsection{New Iterative Method of Subspace Identification}
Our new method iteratively improves the subspace direction by optimizing a function.  
It uses golden-section search (GSS) internally, which given a unimodal function, finds an extremum (minimum or maximum) of the function within a specified interval.
It operates by successively narrowing the range of function values within the interval without using the gradient of the function.
We use WEAT~\cite{CaliskanBrysonNarayanan2017} as the underlying function.
That is given a stereotypical subspace direction $v$, the value $S(v)$ provides the difference in WEAT score \emph{after} applying Linear Projection for subspace $v$.
The smaller the value, the better $v$ captures the bias subspace.

\para{Overfitting, testing, and training data.}
Before we describe our new subspace identification method, it is important to discuss overfitting, testing data, and training data. 
According to its formulation, if WEAT is evaluated on sets of words $A$, $B$, $X$, and $Y$, then it may be unfair to train a subspace on the same words it is evaluated on. 
Otherwise, the trained subspace may not generalize to the vast majority of words not in these sets, i.e., \emph{overfitting}.
For example, if we use definitionally gendered words (e.g., ``man, woman, boy, girl'') in $A$ and $B$, stereotypical occupation words (e.g, ``engineer" and ``receptionist") in $X$ and $Y$, then we should consider defining the subspace using words not in those sets.  
In particular, we explore statistically gendered names (e.g., ``Jack'' for male $M$ and ``Susan'' for female $F$), which define a subspace correlated with the one determined by definitionally gendered words.  

\para{New iterative subspace identification.}
The algorithmic procedure starts with the two-means approach in identifying an initial concept subspace using seed sets $M$ to get mean $m$ and set $F$ to get mean $f$. Initially let $v = m-f$ represent the gender direction.  
Then we iteratively improve the WEAT score $S(v)$ by choosing updated points $m$ in the convex combination of $M$ and $f$ in the convex combination of $F$. 
In each iteration, we fix either $m$ or $f$ and update the other.  
When $f$ is fixed, we cannot use gradient descent to update $m$, since we do not have access to a gradient of the function $S$. 
Rather, we consider moving $m$ towards any point $x \in M$, by setting $m$ to its new location $m_x(\alpha) = (1-\alpha) m + \alpha x$ for $\alpha \in [0,1]$.  
The parameter $\alpha$ represents the fraction towards $x$ from $m$.  
We consider each $x \in M$ in a fixed permutation, and determine how best to move $m$ towards $x$, using GSS to optimize $S(m_x(\alpha))$ as a function of $\alpha$.  
We update $m$ to $m_x(\alpha)$, and then consider the next $x' \in M$ in the permutation, and update $m_x(\alpha)$ to $m_{x'}(\alpha')$ for the best $\alpha'$.  
After completing this permutation, we fix the new location of $m$, and optimize $f$.  These can be alternately optimized. 
We find two rounds of optimizing $m$ and $f$ is sufficient.

While the above procedure is automated, {\tool} is essential in selecting the words used in $M$ and $F$ so we can see how they are correlated with those in $A$ and $B$ respectively.  
It serves to improve seed word selection, a critical step in a debiasing process.

\subsubsection{Evaluation of Subspace Identification Methods}

We evaluate the effectiveness of our new subspace identification method by computing the WEAT~\cite{CaliskanBrysonNarayanan2017} and ECT~\cite{DevPhillips2019} scores using their respective standard datasets, before and after debiasing with Linear Projection. 
For evaluation, so the evaluation of WEAT is not the same as one optimized, we alter the word lists $X$, $Y$ to be stereotypical adjectives
A = \{``strong'', ``intelligent'', ``brave'', ``important''\} and B = \{``pretty'', ``beautiful'', ``shy'', ``homely''\}.  
In ECT, the aggregated male and female words compare the ranks of distances to a list of occupations.
For the classification normal, we debias using linear projection exactly once.
Recall that a desirable ECT is closer to $1$, while a desirable WEAT is closer to $0$.
In addition to our new Iterative Subspace method, we compute the concept direction $v$ using PCA, 2-means, and classification normal.
Paired-PCA was not available since the input words $M$ and $F$ (statistically gendered names) are not paired.
\autoref{tab:subspace-words} shows that for ECT, Iterative Subspace achieves the largest score (nearly the optimal value of $1$).
Note, if we had trained on $A, B$ instead of $M, F$, then 2-means (which gets the second best score) would get the optimal value of $1$.
For WEAT, the iterative method ($0.902$) and the classification normal approach ($0.951$) are both big improvements over 2-means ($1.1$) and PCA ($1.17$).
For the more extensive NLI Test~\cite{dev2019measuring}, we use their large list of gender-occupation bias measuring sentence pairs and record the fraction of sentences classified neutral, a score called Net Neutral~\cite{dev2019measuring}.
The higher the value (closer to 1), the lesser the bias (see~\autoref{sec: NLI test}). In this test, we see a similar improvement with the two methods.
Iterative subspace identification does the best under all three measures.

\begin{table}[]
    \centering
    \begin{tabular}{c|ccc}
    	\hline
    	        Method          &      ECT       &   WEAT (adj)   &    NLI Test    \\ \hline
    	       Baseline         &     0.773      &     1.587      &     0.297      \\
    	          PCA           &     0.905      &      1.17      &     0.346      \\
    	        2-means         &     0.912      &     1.102      &     0.379      \\
    	Classification (1 step) &     0.872      &     0.951      &     0.383      \\
    	  Iterative Subspace    & \textbf{0.966} & \textbf{0.902} & \textbf{0.386} \\ \hline
    \end{tabular}
    \caption{Bias Subspace Selection in Word Embeddings.}
    \label{tab:subspace-words}
    \vspace{-6mm}
\end{table}

\subsection{{\tool} for Merchant data}
\label{sec:merchantcase}
Extensive amounts of transaction data are available to financial organizations.
To utilize such data for applications such as recommendation or fraud detection, it is important to understand the characteristics associated with each unique merchants presented in the data.
Although a lot of the information can be obtained by either calculating summary statistics associated with each merchant (e.g. average price for each transaction) or directly from the merchant (e.g. merchant categories), a general profile containing additional information can be distilled from the data by creating distributed representations using word embedding algorithms~\cite{du2019pcard,wang2019feature,yeh2020towards}.
This is not based on text associated with these merchants, but rather the sequences in which merchants were visited by customers.

The embedding dataset presented in this section is generated from real-world transactions from a global payment company.
It captures payment activities between $70$ million merchants and $260$ million customers from December 1, 2017 to June 30, 2019 in the United States.
The merchant embedding is generated by Word2vec~\cite{HeimerlGleicher2018,du2019pcard}, where each merchant is treated as a word and each customer as a document.

\begin{figure}
	\includegraphics[width=.49\linewidth]{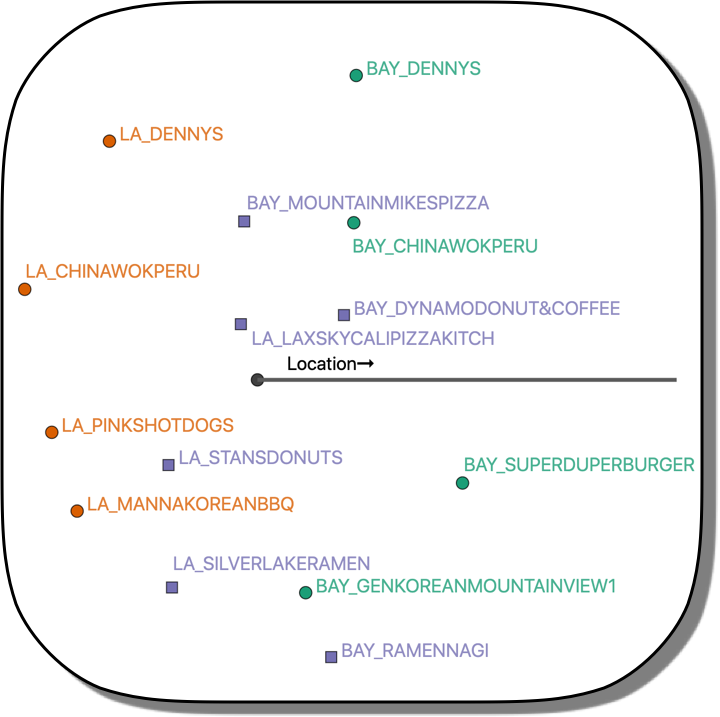}
	\hfill
	\includegraphics[width=.49\linewidth]{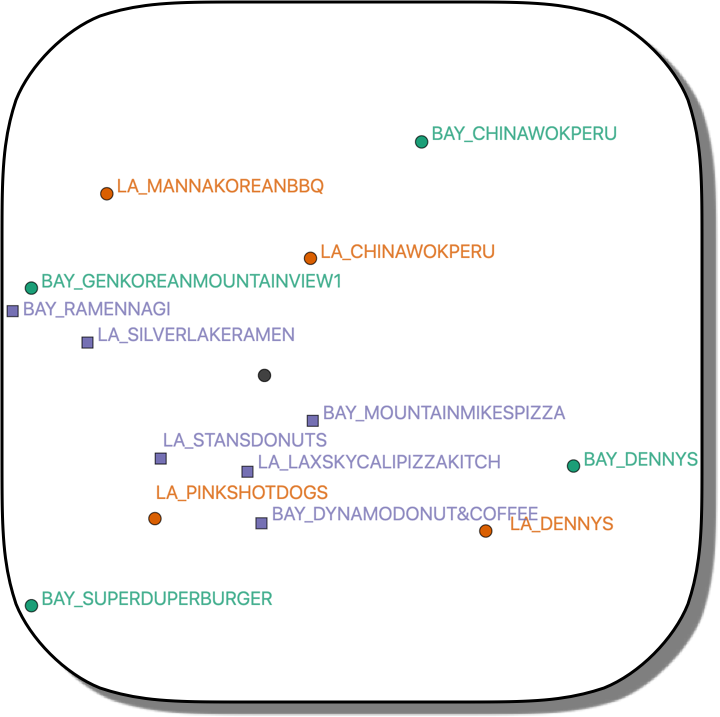}
	\vspace{-2mm}
	\caption{Removing location information from a restaurant embedding using a Linear Projection debiasing technique.
	From left to right, visualizing restaurant embedding before and after Linear Projection.}
	\label{fig:merchant_location}
\end{figure}

With rich information in the merchant embedding, it can be generally applied to many downstream tasks~\cite{patent_fraud, du2019pcard}.
Here we focus on a subset of the merchant embedding, referred to as the \emph{restaurant embedding}, which is extremely important for recommendation system.
By visualizing the merchant embedding dataset through {\tool} (\autoref{fig:merchant_location}), one can find that the distribution of embeddings are significantly dominated by each restaurants' geographical location over other information such as cuisine type.
In other words, geographical location captured by the embedding would interfere the recommendation system when recommending restaurants as a user's ``taste" is usually more associated with other information like cuisine type.
It is essential to tease out the undesired subspaces within the embedding space that represents irrelevant information for better recommendation performance.

In \autoref{fig:merchant_location} (Left), green points are restaurants from the bay area (BAY), orange points are restaurants from the Los Angeles area (LA), and purple points are the evaluation restaurants. 
From the visualization, all the LA restaurants are clustered together in the left part of the figure and all the BAY restaurants are clustered in the right part. The location direction (North-South) are automatically identified by the two-mean method. By using Linear Projection to remove the location subspace, similar restaurants are now clustered together (\autoref{fig:merchant_location}(Right)). 
For both training dataset and evaluation dataset, we can clearly see seven cuisine types here: Chinese, Ramen, Korean BBQ, Dennys, Donuts, Pizza, Burger/Hotdogs. 
Therefore, {\tool} allows for an easy understanding of restaurant information in the merchant embedding and allows users to modify the embedding to adapt to various downstream tasks.

%% file: sec-feedback.tex
\section{Domain Users' Feedback}

We collected feedback from three groups of domain users, after they thoroughly explored {\tool} on different vector embedding datasets.
The first group includes three workshop attendees ($W_1{\sim}W_3$) from AAAI 2021 conference (where we gave the tutorial).
The second group consists of two graduate students ($S_1{\sim}S_2$), majoring in Computer Science.
Their research focuses on studying ML and NLP  techniques, where exploring embedding data is an important part of their daily work. 
The third group includes two researchers ($R_1{\sim}R_2$) of an industrial research lab, who used {\tool} to analyze the merchant embedding data as we have explained in ~\autoref{sec:merchantcase}.

For all three groups of users, we first introduced {\tool} to them and explained the functions of different visual components. 
The users were then given sufficient time (several hours for $W_1{\sim}W_3$ during the workshop, and a couple of days for $S_1{\sim}S_2$ and $R_1{\sim}R_2$). Lastly, feedback was collected from them either in written form and/or remote meetings.

In general, all domain users provide positive feedback on {\tool} in clearly revealing the bias hidden inside embedding data, and intuitively comparing different debiasing algorithms.
For example, $W_1{\sim}W_3$ responded our feedback questions by valuing the easy accessibility of {\tool} and the easy-to-install level was rated between ``extremely-easy'' and ``easy''.
Both $S_1$ and $S_2$ expressed their awareness and worry on the biases in their embedding data.
They reaffirmed the importance of the debiasing problem and  appreciated the interactivity provided by {\tool} in easily disclosing the hidden biases.
Additionally, we also had thorough discussions with them on other design choices in presenting the clustering results with different dimensionality reduction algorithms.
$R_1$ commented that {\tool} concretized the bias mitigation process in her mind through smooth animations and it demonstrated the process in a more intuitive manner, which significantly helped her understand the merchant embedding data.
Compared to the traditional way of manually comparing the embeddings before and after debiasing, {\tool} is more convenient, interactive, and user-friendly.

There are also several suggested improvements provided by the users.
First, $S_2$ discussed the importance of debaising in language translation tasks to prevent the propagation of bias from one language to the other.
He suggested concurrently analyzing multiple sets of embeddings in the same projection space to investigate and relate the biases from individual sets.
Adding contexts for biases is another interesting comments, as the biases from one language may not be biases in the other.
Second, $S_1{\sim}S2$, as well as the workshop attendees, were impressed by {\tool} integrating so many debiasing algorithms.
However, some of them (e.g., Hard Debiasing) were not very familiar to them.
$S_1$ recommended adding short video clips, or links to some algorithm explanations to briefly explain different debiasing techniques.
Lastly, some direct side-by-side comparisons for different debiasing algorithms were also recommended by the users (without considering the space usage).
The feedback provides promising further directions for us to explore.

%% file: sec-conclusions.tex
\section{Conclusions and Discussions}
\label{sec:conclusions}

This paper presents VERB, a new tool for visualizing, interacting with, and teaching embedded representations of data.  
It is especially useful at allowing users to interact and observe the effects of debiasing word vector embeddings -- an essential component of most NLP tasks, and critical for ensuring fairness.  
VERB is distinct from previous visualization tools for high-dimensional vectorized data in that it is used to modify and improve the vectors, before they are used in downstream tasks, not just explore or inspect them.  

The {\tool} visual tool is useful in other ways, it allows one to find new types of bias, it identifies the importance of subspace determination in addition to the debiasing mechanisms -- leading to an improved method, and it is demonstrated to extend to other forms of data including merchants from a large payment transaction company.  

A key challenge VERB addresses is how to show high-dimensional embedded representions informatively when only a 2-dimensional view is visually available.
It relies on displaying this vectorized data as projected views, via \emph{linear} projections.
This linearity is essential since the debiasing and modification relies on identification and movement of data along linear subspaces.  Non-linear methods like t-SNE or ISOMAP would distort these linear objects.  
Moreover, these identified concept subspaces are essential to defining these views.
These identified subspaces define the $x$- and possibly $y$-axis of the view, where users can clearly see the amount of contribution individual words have along that subspace, without concerning the possible distortions introduced by a skewed projection.  

Overall, VERB is a simple and easy-to-use interface for understanding and acting on a wide variety of vectorized representations.
It demystifies, and provides a powerful way to interact with and debias these representations, towards interpretable and fair ML that operates on these representations.